\documentclass{article}

\usepackage{arxiv}

\usepackage[utf8]{inputenc} % allow utf-8 input
\usepackage[T1]{fontenc}    % use 8-bit T1 fonts
\usepackage{hyperref}       % hyperlinks
\usepackage{url}            % simple URL typesetting
\usepackage{booktabs}       % professional-quality tables
\usepackage{amsfonts}       % blackboard math symbols
\usepackage{nicefrac}       % compact symbols for 1/2, etc.
\usepackage{microtype}      % microtypography
\usepackage{lipsum}		% Can be removed after putting your text content
\usepackage{graphicx}
\usepackage{natbib}
\usepackage{doi}

\usepackage{xcolor}
\usepackage{amssymb}
\usepackage{bm}
\usepackage{subfigure}
\usepackage{amsmath}
\usepackage[most]{tcolorbox}

\title{Rethinking Reinforcement fine-tuning of LLMs:  
A Multi-armed Bandit Learning Perspective}

\author{
Xiao Hu\textsuperscript{1}\quad
Hong Xie\textsuperscript{1}\thanks{Corresponding author}\quad
Tao Tan\textsuperscript{1}\quad
Defu Lian\textsuperscript{1}\quad
Jianyu Han\textsuperscript{2}\quad
\\[4pt]
\textsuperscript{1}University of Science and Technology of China\quad
\textsuperscript{2}IFlyTek (China)
}

\date{}

\hypersetup{
pdftitle={Rethinking Reinforcement fine-tuning of LLMs:  
A Multi-armed Bandit Learning Perspective},
pdfauthor={Xiao Hu, Hong Xie, Tao Tan, Defu Lian, Jianyu Han},
}

\begin{document}
\maketitle

\begin{abstract}
A large number of heuristics have been proposed to optimize the 
reinforcement fine-tuning of LLMs.  
However, inconsistent claims are made from time to time, 
making this area elusive.  
Reflecting on this situation, 
two fundamental questions still lack a clear understanding: 
1) what is the role of each optimizing choice? 
2) which ones are the bottlenecks? 
This paper aims to shed light on them, 
and it faces the challenge of several entangled confounding factors 
in the fine-tuning process.   
To tackle this challenge, 
we propose a bottom-up experiment pipeline.  
The bottom layer is composed of a minimalist configuration: 
one training data, one rollout per round and 
the reward directly serve as the learning signal without advantage function design.  
This minimalist configuration connects to 
multi-armed bandit learning 
with extremely large discrete action space, 
which offers theories to corroborate the experiment findings.   
The up procedure of the experiment pipeline 
expanding the minimalist configuration layer by layer, 
examining the role of each design choice.   
Experimental results on three LLMs 
and two reasoning datasets not only reveal new understanding 
of the design choice but also yield essential insights to 
shape the area.     
\end{abstract}

% keywords can be removed
% \keywords{First keyword \and Second keyword \and More}

\section{Introduction}
Reinforcement fine-tuning of LLMs is drawing extensive attention from both academia and industry \cite{Wang2025,Zhang2025a}.  Various aspects of it were optimized, aiming to unlock the full potential of LLMs.  Tracing back, the literature was started by directly applying policy-based reinforcement learning algorithms to fine-tune LLMs \cite{PPO17,InstructGPT22,ChatGPT22}.  Critic-free policy algorithms \cite{Rafailov2023,GRPO-DeepSeekMath24} definitely established an important milestone.  Various aspects of the critic-free policy algorithms have been optimized, such as credit assignment mechanism \cite{Kazemnejad2025}, entropy regularization \cite{Wang2025}, advantage function \cite{Li2024}, gradient optimization \cite{DAPO25}, etc. 
More recently, scaling the reinforcement fine-tuning in terms of increasing batch size, number of rollout, etc., are shown to achieve performance from a new dimension  \cite{Scaling_Law_for_LLM24,ScaleRL25,Depth-Breadth_Synergy25,Act_Only_Pay25,GRPO-MA25}. The literature is vast and expands quickly, making it impossible for us to enumerate them all, and thus we name only a few typical works as above.  

However, the reinforcement fine-tuning area is not truly flourishing despite the vast number of papers, but instead the area is undergoing various critical challenges.  First,  inconsistent claims are made from time to time.  Take the effect of policy entropy as an example \cite{Wang2025}.  It was claimed in several studies that higher entropy promotes exploration and improves performance\cite{DAPO25,Entropy-Perspective25,GTPO&GRPO-S25}.  The opposite side was also claimed, i.e., lower entropy leads to better performance \cite{Entropy-Minimization25}.  It was also claimed that reducing entropy does not necessarily improve model performance\cite{ExplorVSExploi25}.  Furthermore, random rewards are even shown to improve the test accuracy of base models like Qwen \cite{SpuriousRewards25}.  Lastly, the generalization capability of reinforcement fine-tuning is still under debate \cite{Yue2025,Chu2025}.  

Reflecting on the above situation, one underlying reason for the illusion is that two fundamental questions still lack a clear understanding: 
1) what is the role of each optimizing choice? 
2) which ones are the bottlenecks?  
Several empirical attempts studied the role of reinforcement learning from the perspective of information theory \cite{Swamy2025}, the role of data influence \cite{Tan2025}, the role of generalization capability \cite{Jin2025}, etc. Furthermore, several theoretical attempts studied the impact of base model \cite{Dylan2025},  the limits of outcome-based reward \cite{Chen2025}, sample efficiency \cite{Xie2025,Zhao2025}, etc. Though a number of novel insights are yielded, there is still some distance to satisfactory answers.  The empirical line of study is calling for better approaches to combating the confounding factors that lead to spurious findings.  The theoretical line falls short in practice,  due to the fact that some important factors of LLMs, like their internal mechanism, are notoriously difficult to model.  

It is challenging to answer the aforementioned questions.  The LLM base model stores various knowledge and the black box knowledge conflict mechanism creates many confounding factors  that make the learning and generalization hard to interpret and relate.  Besides, in the training process, many factors like the data, the reward signal, learning rate, etc.,  influence the learning dynamics as well as generalization behavior. What makes the problem more complicated is that these factors are entangled together.  

To tackle the above challenges, this paper rethinks the reinforcement fine-tuning in a bottom-up manner.  
The bottom layer is composed of a minimalist configuration: 
one training data, one rollout per round and 
the reward directly serve as the learning signal without advantage function design.  
This minimalist configuration connects to 
multi-armed bandit learning 
with extremely large discrete action space \cite{Lattimore2020}, 
which offers theories to corroborate the experiment findings.   
More interestingly, 
it is a simplified multi-armed bandit learning model in the sense that the reward is deterministic and more importantly, each reward indicates the optimality of the action (or correctness of the answer).  
The up procedure of the experiment pipeline 
expanding the minimalist configuration layer by layer, 
examining the role of each design choice.   
Experimental results on three LLMs 
and two reasoning datasets not only reveal new understanding 
of the design choice but also yield essential insights to 
shape the reinforcement fine-tuning area. 
%show that under the baseline case   
The contributions are highlighted as follows.

\subsection{Contribution}

\paragraph{A bottom-up experiment pipeline.}
Inspired by multi-armed bandit learning, 
a bottom-up experiment pipeline is proposed, 
which disentangles various factors that influence the reinforcement fine-tuning.  
It is generic and can be applied to analyze other scenarios including agent fine-tuning.  

\paragraph{Critical understanding of reinforcement fine-tuning.}
Critical analysis the training and generalization dynamics, revealed that: 
1) under the minimalist configuration, all three base models can achieve Pass@1 score of 1 
on the training data, and the Pass@1 over the test data can be improved by as high as 
{\color{blue}0.5}; 
2) varying the difficulty levels of training data, both the training and testing curves vary slightly, unless the data is extremely difficult (Pass@1 less than 0.05), which leads to drastic Pass@1 drop over the test data;  3) when the training data is of moderate difficulty level (larger than 0.1), increasing the number of rollouts increases the learning speed, but it does not yield an improvement in the test Pass@1.
4) the negative reward metric, i.e., $\{0,-1\}$,  leads to diverging training and drastic Pass@1 drop over the test data;
5) the GRPO-like advantage function does not make a difference on training and generalization, 
unless the training data is extremely difficult, where advantage function reduces the risk of drastic Pass@1 drop over the test data; 
6) the OLMo model exhibits different behavior than Qwen and LLaMA in that it learns well but it does not generalize. 

\paragraph{Insights for shaping the area.}
The above experimental findings suggest that many factors like advantage function, number of rollouts, data selection, etc., are not as critical as previous studies claim.  The reinforcement fine-tuning should pay more attention to understanding the corner cases (such as extremely difficult data) in which the model does not generalize, and possibly designing better algorithm to enable generalization.  We envision that for moderate cases, i.e., a good base model and moderate difficult data, finer design of advantage function, or sampling strategy, etc., would not contribute significant improvement, in such cases simple ones would do well enough.    

\section{Related Work}
\paragraph{Connections to reinforcement learning. }
Though the literature on reinforcement fine-tuning is large and most of it follows the principles of reinforcement learning, the connection between reinforcement fine-tuning and reinforcement learning is still unclear \cite{Wang2025,Zhang2025a}.  Several empirical examined the role of reinforcement learning from the lens of likelihood  \cite{Swamy2025}, out of distribution generalization capability \cite{Jin2025}, etc. Furthermore, several theoretical attempts studied the impact of base model on policy learning \cite{Dylan2025},  limits of outcome-based reward \cite{Chen2025}, sample efficiency \cite{Xie2025,Zhao2025}, etc.  A number of novel insights are yielded, but there is still a long way to go, since the gap between theory and practice is large.  

\paragraph{Advantage optimization.}
Recent critic-free RLVR/RLFT methods, such as GRPO \cite{GRPO-DeepSeekMath24} and its variants, generally regard advantage design as a key mechanism.  Accordingly, prior work has mainly focused on baseline construction and normalization strategies to improve convergence efficiency and mitigate training instability \cite{REINFORCE++25,GRPO-Norm25,QAE25,DAPO25,R1ZeroCrit25}.  This work shows that the advantage function does improve the generalization as the literature claims.  It implies that the role of advantage function should be re-examined.  

\paragraph{Reward function.}  
In terms of reward design, some studies reported that penalizing incorrect outputs with negative rewards can improve pass@k under certain settings, typically by suppressing erroneous modes and redistributing probability mass~\cite{NegReinf25}.  However, some analyses reveal that the effectiveness of rewards may be highly sensitive to model priors and algorithmic biases, leading to inconsistent outcomes across models and tasks and thus requiring cautious interpretation~\cite{SpuriousRewards25}.  Against this backdrop, our work conducts systematic experiments to further characterize the boundaries of these optimization signals in terms of convergence speed and generalization behavior.  

\paragraph{Task difficulty.} 
The performance of reinforcement learning is closely tied to data quality~\cite{LIMO25}, and existing studies often associate data quality with task difficulty. Some works argue that moderately difficult examples are more effective in maintaining stable and informative training signals, thereby improving data efficiency~\cite{ImproData25}. In contrast, other studies emphasize that hard examples are particularly beneficial for performance gains and out-of-distribution generalization~\cite{HardExamples25,DataAttributes25}. In addition, prior work highlights that long-tail examples remain a primary bottleneck for model generalization~\cite{LLmStruggle23}. In contrast, our experimental results indicate that, for non–long-tail examples, task difficulty has a relatively limited impact on generalization performance, whereas under long-tail settings, the interaction between difficulty and training dynamics becomes considerably more complex, revealing phenomena that merit  investigation.  

\paragraph{The role of base model.}  
Various empirical studies revealed that the effectiveness of RLVR is largely constrained by the pretraining characteristics of the base model and its alignment with the target task~\cite{OctoThinker25,EchoChamber25}.  Moreover, some works point out that certain base models exhibit distinctive behaviors in reward responses or optimization dynamics, and neglecting such differences in algorithm design or experimental analysis may lead to misleading or non-generalizable conclusions~\cite{SpuriousRewards25}.  From a theoretical perspective, the impact of base model on sample complexity of policy learning is proved \cite{Dylan2025}.  Our work moves one step further by revealing that the alignment is not enough, and that extreme difficult questions makes a huge difference in learning and generalization.  

\paragraph{Scaling reinforcement fine-tuning.}  
Recently, a number of works aim at scaling the reinforcement fine-tuning, in terms of increasing batch size, number of rollout, etc, showing improvement in generalization \cite{Scaling_Law_for_LLM24,ScaleRL25,Depth-Breadth_Synergy25,Act_Only_Pay25,GRPO-MA25}. Our work shows that increasing the number of rollouts per round does not improve generalization.  This inconsistency between literature highlights the existence of hidden factors worth further digging out.  

\section{Background and Problem Statement}
\subsection{Reinforcement Fine-tuning}
This paper considers reinforcement fine-tuning of LLMs 
with verifiable rewards where the reward is outcome-level.  
Let $q$ denote a query and $\mathcal{O}$ denote the response space.  
The base model is denoted by a policy $\pi_{\theta}$, which prescribes 
a probability distribution over $\mathcal{O}$ 
for each query $\pi_{\theta} ( \cdot| q)$, namely 
$\sum_{\bm{y} \in \mathcal{O}} \pi_{\theta}( \cdot| q) = 1$ 
and $\pi_{\theta}( \cdot| q) \in [0,1]$, where $\theta$ denotes the parameters of the 
base model.  
We consider a binary reward metric $\{0,1\}$ with 
0/1 indicating whether a response is incorrect/correct.  
The training dataset contains only one query 
denoted by $\mathcal{D}_{train} =\{q\}$ 
and the test dataset contains $N \in \mathbb{N}_+$ queries 
denoted by $\mathcal{D}_{test} =\{q_1, \ldots, q_N\}$.  
Throughout this paper, the Pass@1 serves as the accuracy metric 
for evaluating the LLM.  

\subsection{Connections to Multi-armed Bandit Learning}  
Multi-armed bandits are a sequential decision framework to study the exploration vs. exploitation trade-offs \cite{Lattimore2020}.  It is composed of a finite number of arms.  Each arm is associated with an unknown reward function, which is usually modeled by a random variable.   In each round, one arm can be pulled and  a reward is generated from the pulled arm.  The optimal policy is the one that pulls the arm with the largest reward mean.  The objective is to learn the optimal policy through interactions with the environment.  One typical variant of multi-armed bandits is multi-play multi-armed bandits, which allows the agent to pull multiple arms in each round.  

Reinforcement fine-tuning with one training data connects to multi-armed bandits as follows.  Each possible response can be modeled by an arm.  Generating a response corresponds to pulling an arm and  the outcome level reward corresponds to the reward generated by an arm.  The correct response corresponds to the optimal arm.  Note that multiple optimal arms are allowed in the multi-armed bandits framework. Reinforcement fine-tuning simplifies reward function of multi-armed bandit learning in the sense that the reward is deterministic  and more importantly each reward indicates the optimality of the action (or correctness of the answer).  It complicates the multi-armed bandit learning in three aspects: 
(1) the black box base model; 
(2) a huge action space with correlated actions. 
(3) the goal of generalization performance. 

Multi-armed bandits are simpler than reinforcement learning in that they do not have state transition.  In the literature, various simple and efficient algorithms are developed, where the reward usually directly serves as the signal for policy learning \cite{Lattimore2020}.  Furthermore, theories indicate that pulling one arm per round (corresponds to one rollout per training round) can enable optimal policy learning already.     

\subsection{Problem Statement}
\label{sec:problem}
The above connection to multi-armed bandit learning  
inspires us to construct the minimalist configuration for reinforcement  fine-tuning, 
which is summarized as follows: 
\begin{itemize}

\item 
One training data  $\mathcal{D}_{train} =\{q\}$.

\item 
One rollout in each training round.  

\item 
The outcome reward directly serves as the training signal without 
any advantage design. 

\end{itemize}

Formally, the rollout generated in the $i$-th training round is denoted by $\bm{o}_i$.  
The outcome reward of  $\bm{o}_i$ is denoted by $R_i \in \{0,1\}$.  
The training objective function is a simplification of GRPO \cite{GRPO-DeepSeekMath24}: 
\[
\begin{aligned}
\mathcal{J}(\theta)
{=}
& 
\min \Big[
r_{i,t}(\theta) R_i,\;
\text{clip}\!\left(r_{i,t}(\theta), 1{-}\varepsilon, 1{+}\varepsilon\right) R_i
\Big]
\\
&
{-} \beta  \mathbb{D}_{\text{KL}}\!\left[\pi_{\theta} \,\|\, \pi_{\text{ref}}\right].
\end{aligned}
\]
Here, the $r_{i,t}(\theta), \varepsilon, \beta$, etc., 
have the same meaning as they do in GRPO \cite{GRPO-DeepSeekMath24}.  
Centering around this minimalist configuration,  
we aim to design a generic experimental pipeline 
to carefully examine the role of each design choice. 

\section{Experiments}
\subsection{Experiment Setting}
We consider three instruction-tuned large language models: LLaMA-3.2-1B-Instruct~\cite{LlaMA3}, OLMo-2-0425-1B-Instruct~\cite{OLMo}, and Qwen2.5-1.5B-Instruct~\cite{Qwen25}.  
We adopt two open-source mathematical reasoning benchmarks, i.e., 
MATH~\cite{MATH21} and GSM8K~\cite{GSM8K21}, for training and evaluation. Following a single-example training protocol, each training dataset consists of a single problem instance sampled from either MATH or GSM8K. To study the effect of problem difficulty, we construct multiple such single-example training sets by selecting problems with varying difficulty levels. 
Specifically, the selected MATH problems are denoted as $\pi_1, \pi_2, \ldots, \pi_7$, and the selected GSM8K problems are denoted as $\phi_1$. Our implementation employs the VeRL framework~\cite{VeRL24}.
Due to page limit, 
the exact training problems and their ground-truth answers, 
and more details on the experiment setting  are provided in appendix.  

\subsection{Exp1: The Minimalist Configuration}
\label{sec:Exp1}
In this experiment, we study the minimalist configuration stated in Section \ref{sec:problem}.  
It would serve as the baseline for later experiment comparison 
for insights yielding.  

\begin{figure}[!htb]
\centering
\vspace{-0.08in}

% -------- Figure A: LLaMA --------
\begin{minipage}[t]{0.49\linewidth}
\centering
\subfigure[GSM8K dataset]{
  \includegraphics[width=0.95\linewidth]{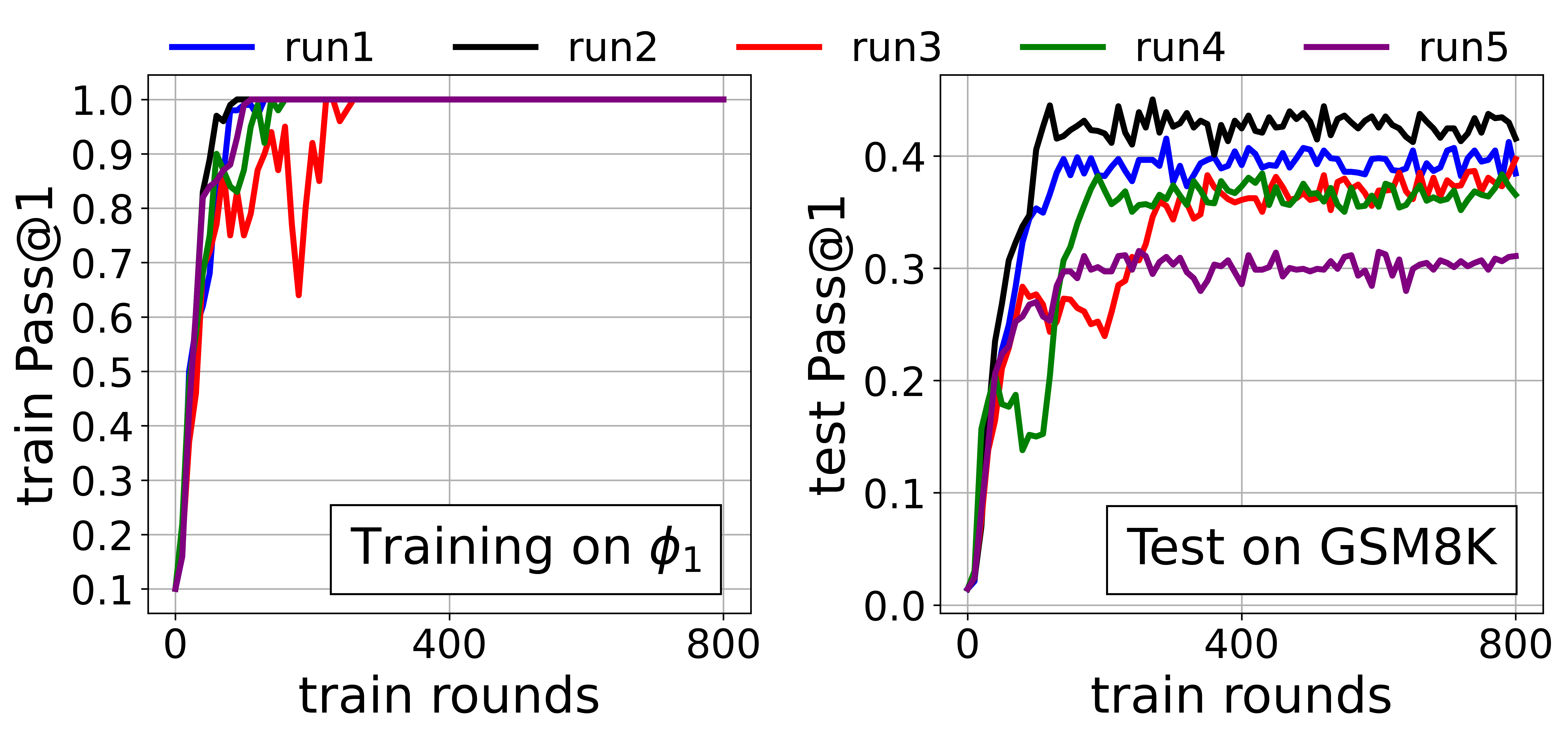}
  \label{gsm8k-rollout-1-llama-phi1-Exp1}
}\\[-0.08in]
\subfigure[MATH dataset]{
  \includegraphics[width=0.95\linewidth]{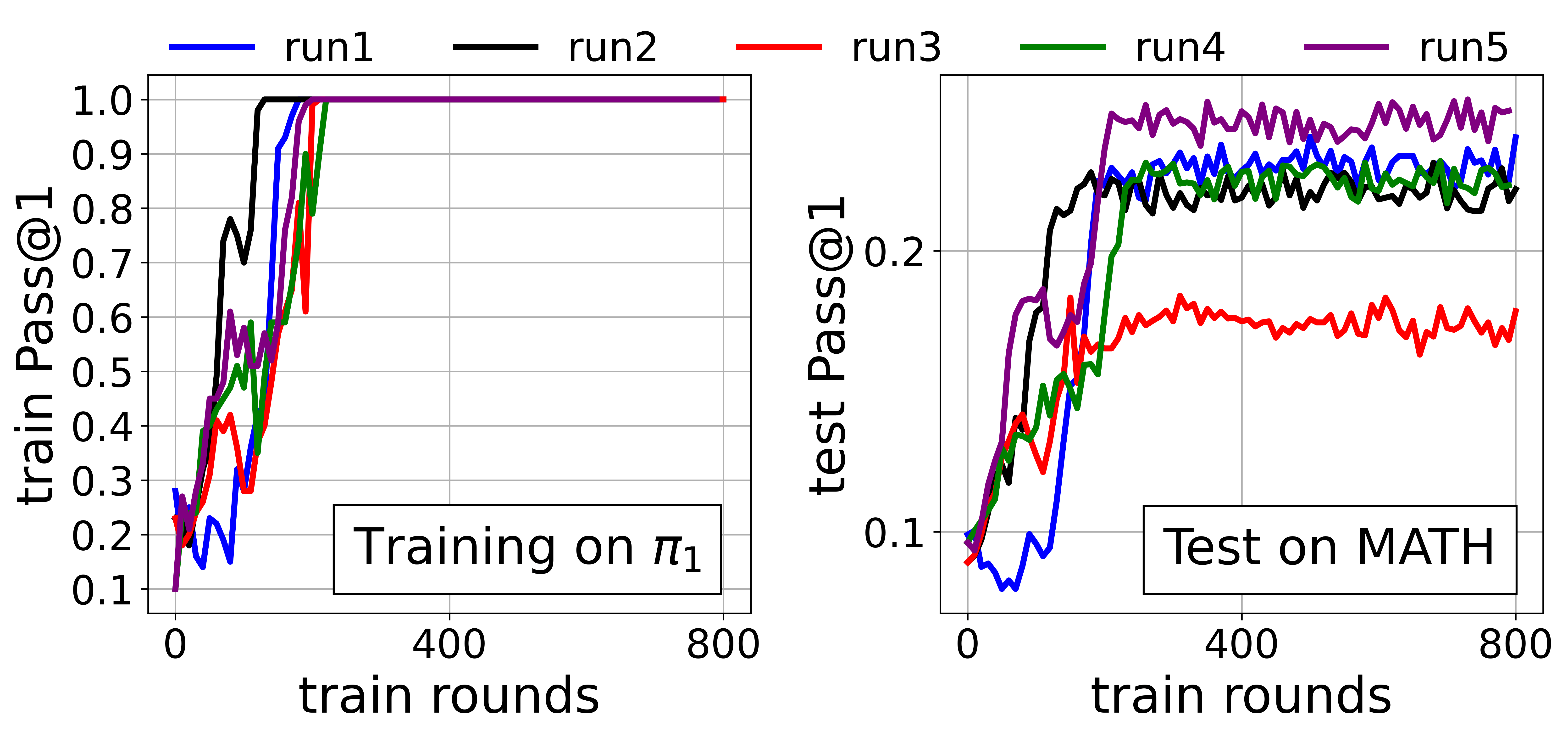}
  \label{rollout-1-llama-hard-pi1-Exp1}
}
\caption{LLaMA fine-tuning under minimalist configuration.}
\label{fig:LLaMA-minimalist-hard}
\end{minipage}
\hfill
% -------- Figure B: Qwen --------
\begin{minipage}[t]{0.49\linewidth}
\centering
\subfigure[GSM8K dataset]{
  \includegraphics[width=0.95\linewidth]{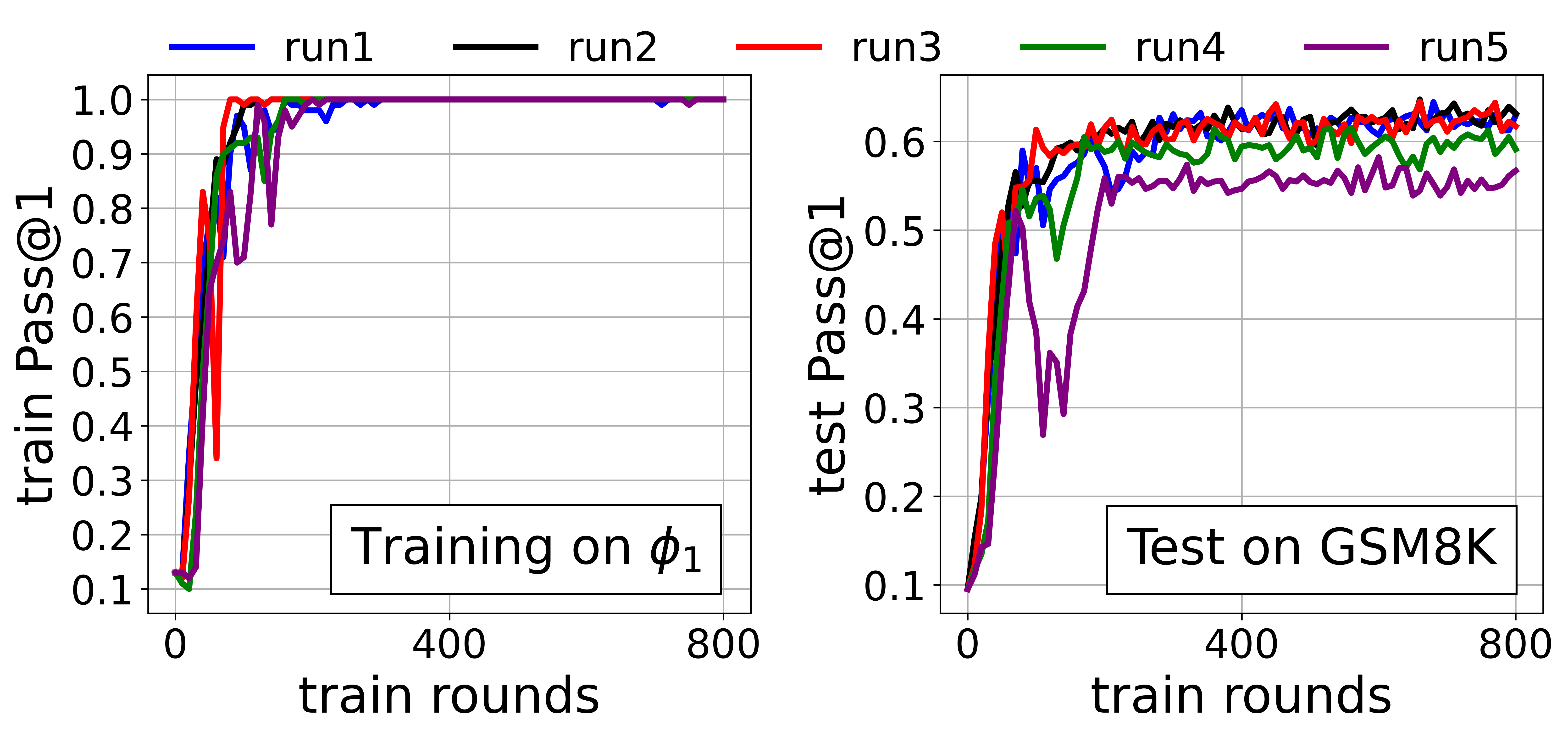}
  \label{gsm8k-rollout-1-qwen15-phi1-Exp1}
}\\[-0.08in]
\subfigure[MATH dataset]{
  \includegraphics[width=0.95\linewidth]{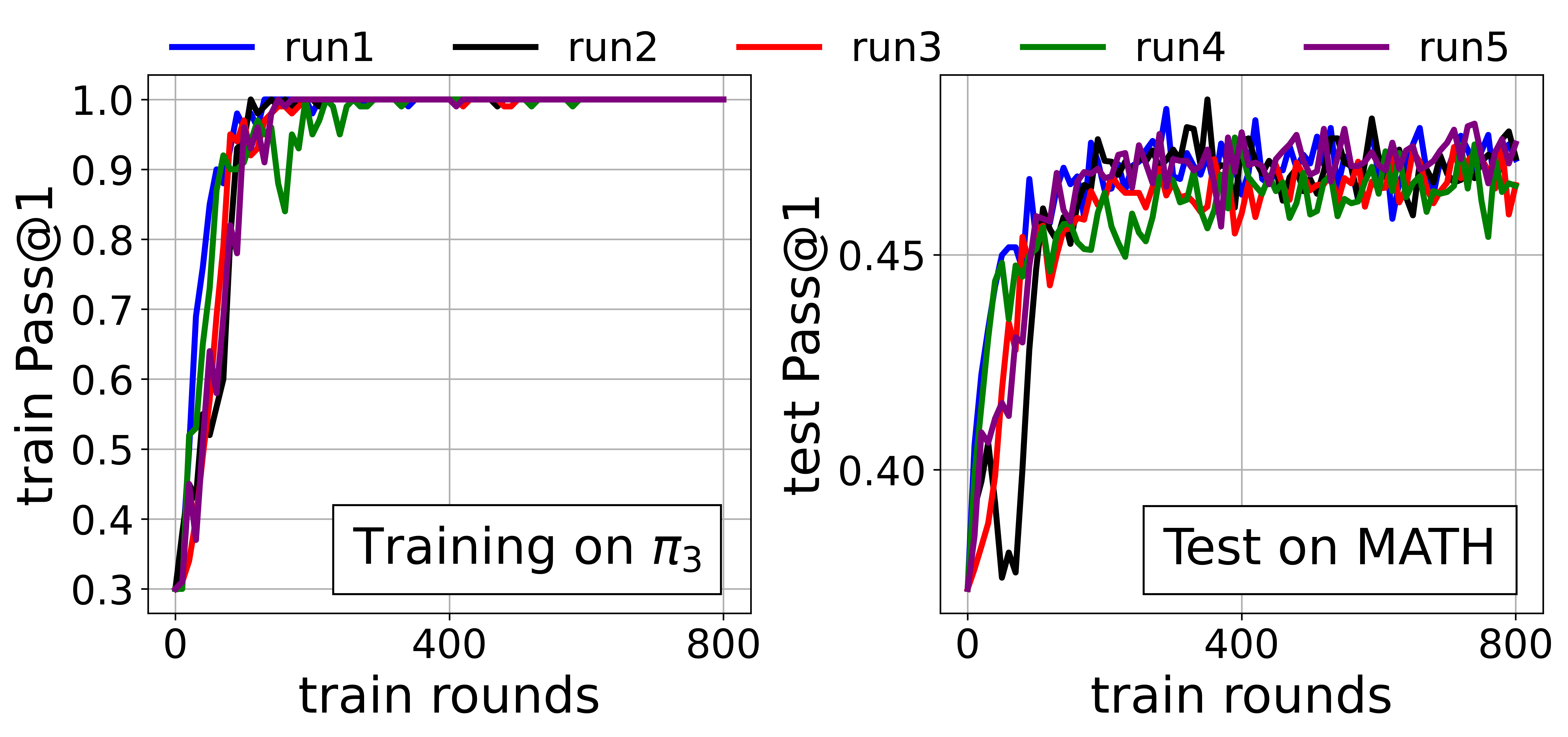}
  \label{rollout-1-qwen15-hard-pi3-Exp1}
}
\caption{Qwen fine-tuning under minimalist configuration.}
\label{fig:QWEN-minimalist-hard}
\end{minipage}

\vspace{-0.08in}
\end{figure}

\paragraph{Training dynamics.} 
We first instantiate the minimalist configuration with the instance $\phi_1$ selected from GSM8K.  
The left column of Figure \ref{gsm8k-rollout-1-llama-phi1-Exp1} shows the training dynamics 
of fine-tuning LLaMA under this setting.  
The horizontal axis represents the number of training rounds, 
while the vertical axis represents the train Pass@1 evaluated on the training data.  
There are five curves corresponding to five independent runs of the same training setting.
As the number of rounds increases from one to around 200, the Pass@1 increases from around 0.1 to 1. Furthermore, it plateaus when the number of training rounds is further increased.  
This implies that LLaMA is able to generate correct answer with probability 1 for the training data 
using around 200 rounds.  
In multi-armed bandit learning tongue, it learns the optimal policy.  
Furthermore, before the Pass@1 converges to 1, 
there are certain variations among five curves, 
showing some certain randomness in the training dynamics.  
The left column of  Figure ~\ref{rollout-1-llama-hard-pi1-Exp1} shows that LLaMA has 
similar learning dynamics patterns 
when when the training dataset is changed to the instance $\pi_1$ from MATH.  
In the left columns of Figure \ref{fig:QWEN-minimalist-hard} show that the learning dynamics curves of Qwen have similar patterns to LLaMA, 
except that the Pass@1 evaluated on the test data can be improved by as high as 0.5.  
We also conducted more experiments on training data with different difficulty levels, 
and they show similar fine-tuning dynamics patterns.  
Due to space limitations, we present them in appendix. 
These results further corroborate that the learning dynamics pattern is essential.   
These observations are aligned with multi-armed bandit learning, in that 
the optimal policy can be learned without advantage design.  

\paragraph{Generalization dynamics.} 
The right column of Figure \ref{gsm8k-rollout-1-llama-phi1-Exp1} shows 
five generalization dynamics curves of LLaMA, each associated with one training dynamics curve from the left column.   
The horizontal axis represents the number of fine-tuning rounds, 
while the vertical axis represents the Pass@1 evaluated over 
the entire MATH test set.  
As the number of fine-tuning rounds increases 
from one to around 200, the Pass@1 increases from around 0.02 to around 0.4.  
It becomes nearly flat when 
the number of fine-tuning rounds is further increased.  
This shows that using only one training data instance, 
the Pass@1 can be increased by up to around 0.4--a dramatic improvement.  
The variation among five generalization dynamics curves is around 0.1, reflecting the randomness nature of the generalization dynamics.  
From Figure \ref{gsm8k-rollout-1-llama-phi1-Exp1}, 
it is interesting to observe that the training and generalization curves share the same flat region, 
and converging faster in training does not warrant higher generalization capability.  
The right column of  Figure ~\ref{rollout-1-llama-hard-pi1-Exp1} shows that LLaMA has 
similar generalization dynamics patterns 
when the training dataset is changed to the instance $\pi_1$ from MATH.  
In the right columns of  Figure \ref{fig:QWEN-minimalist-hard}, one can observe that 
the generalization dynamics curves of Qwen have similar patterns to LLaMA.  
Additional experiments with different difficulty levels in appendix show similar generalization dynamics patterns, further supporting that the generalization dynamics pattern is essential.

\paragraph{Key insights.}  
Under the minimalist configuration, 
the optimal policy can be learned, 
and the Pass@1 evaluated on the test dataset 
can be improved by as high as 0.5.  

\subsection{Exp2: Impact of Advantage Function}
\label{sec:advF}
To study the impact of advantage function,  
we only modify one component of the configuration in Section \ref{sec:Exp1}, 
i.e., replacing the reward with a GRPO-type advantage.  
To achieve this, we generate multiple rollouts to obtain multiple rewards 
(in our experiment,we generate 4 rollouts).  
Note that only the advantage of the first rollout is used to fine-tune the model.  
Formally, the GRPO advantage formula is calculated as 
$
adv = (r_1 - mean\{r_1, r_2, ..., r_G\}) / std\{r_1, r_2, ..., r_G\},
$
where $G$ denotes the number of rollouts.  

\begin{figure}[!htb]
\centering
\vspace{-0.08in}

% -------- Figure A: LLaMA --------
\begin{minipage}[t]{0.49\linewidth}
\centering
\subfigure[LLaMA with GRPO advantage]{
  \includegraphics[width=0.95\linewidth]{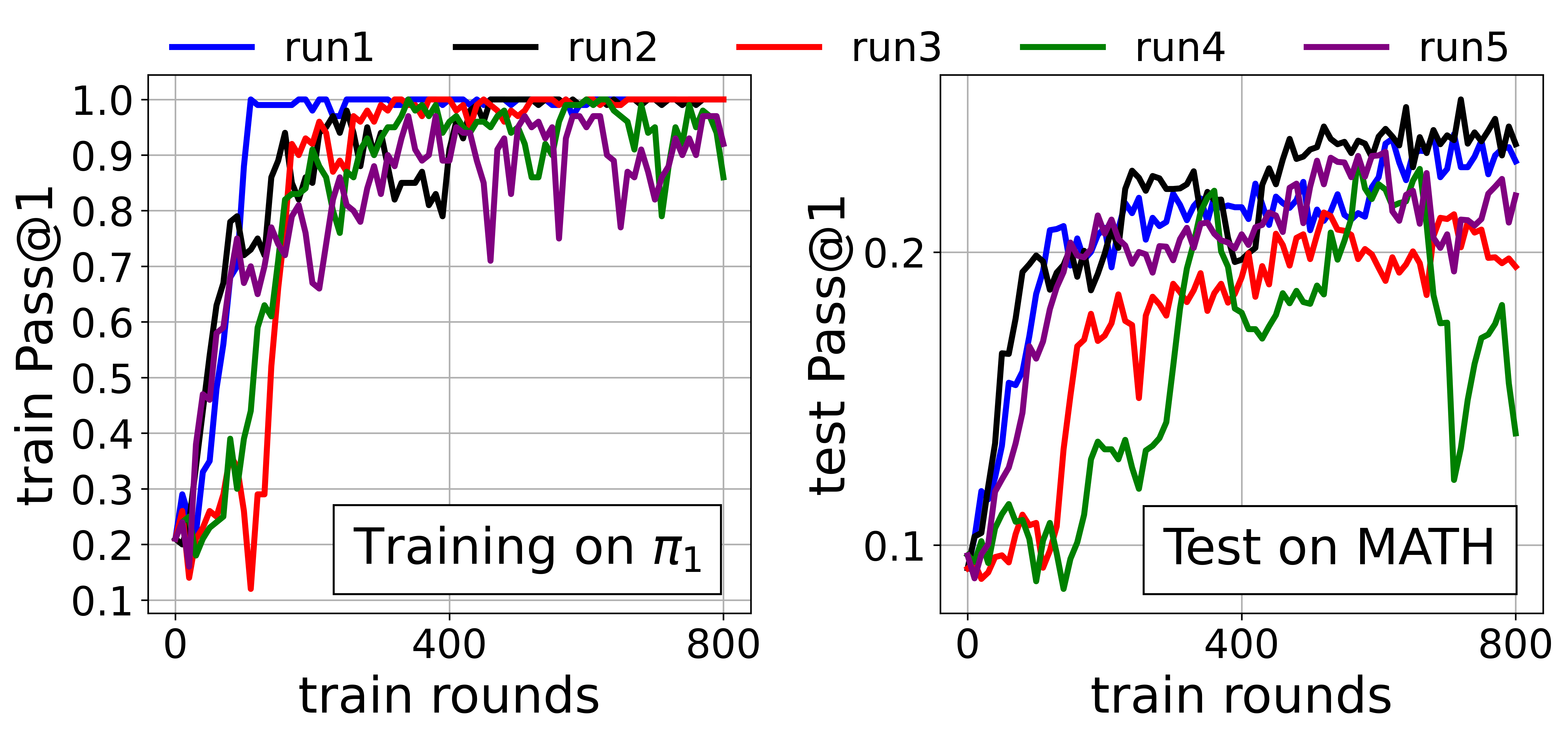}
  \label{only-one-llama-pi1-Exp2}
}\\[-0.08in]
\subfigure[Qwen with GRPO advantage]{
  \includegraphics[width=0.95\linewidth]{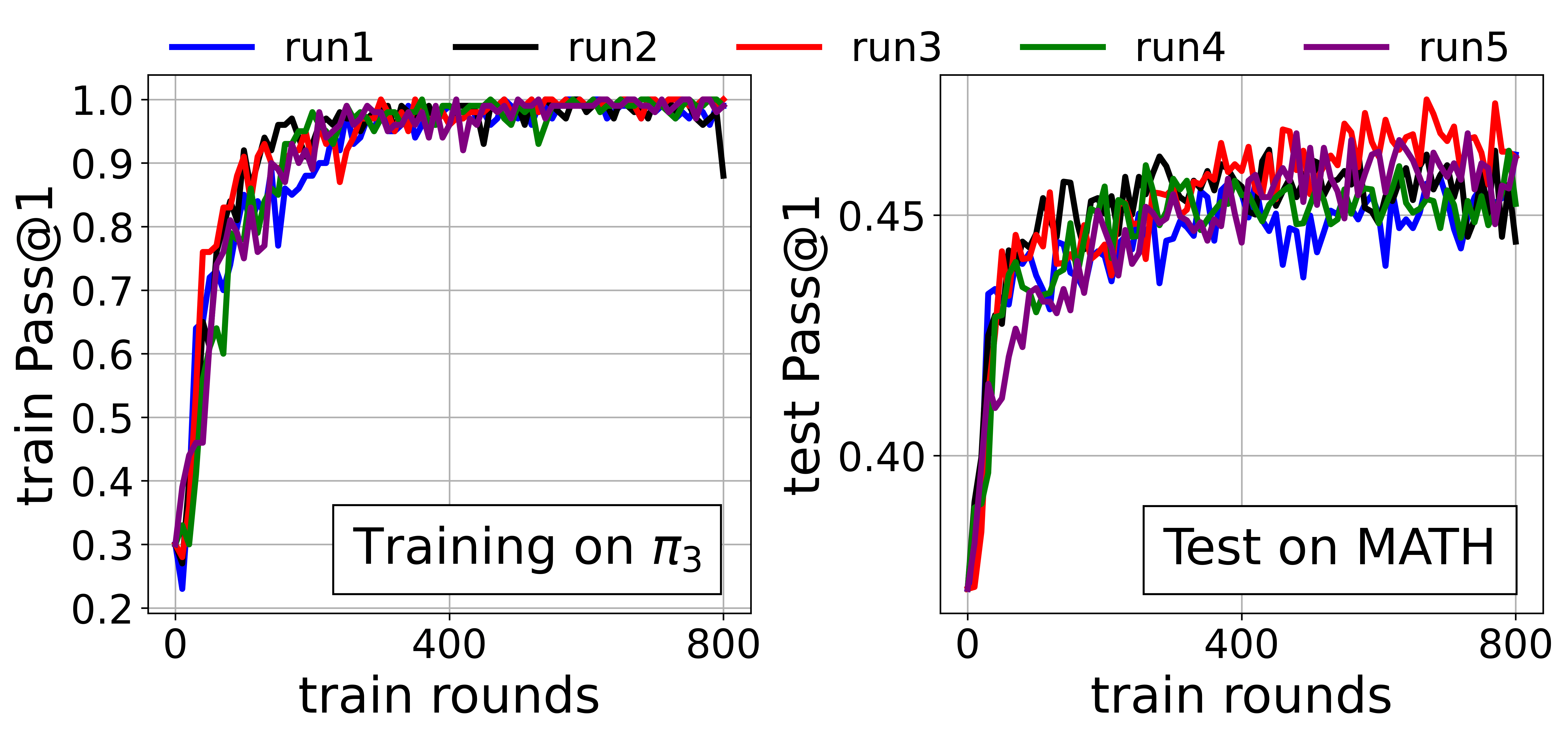}
  \label{only-one-qwen15-pi3-Exp2}
}
\caption{Impact of advantage function (Math dataset).}
\label{fig:ImpactAdvantage-hard}
\end{minipage}
\hfill
% -------- Figure B: Qwen --------
\begin{minipage}[t]{0.49\linewidth}
\centering
\subfigure[LLaMA with 4 rollouts]{
  \includegraphics[width=0.95\linewidth]{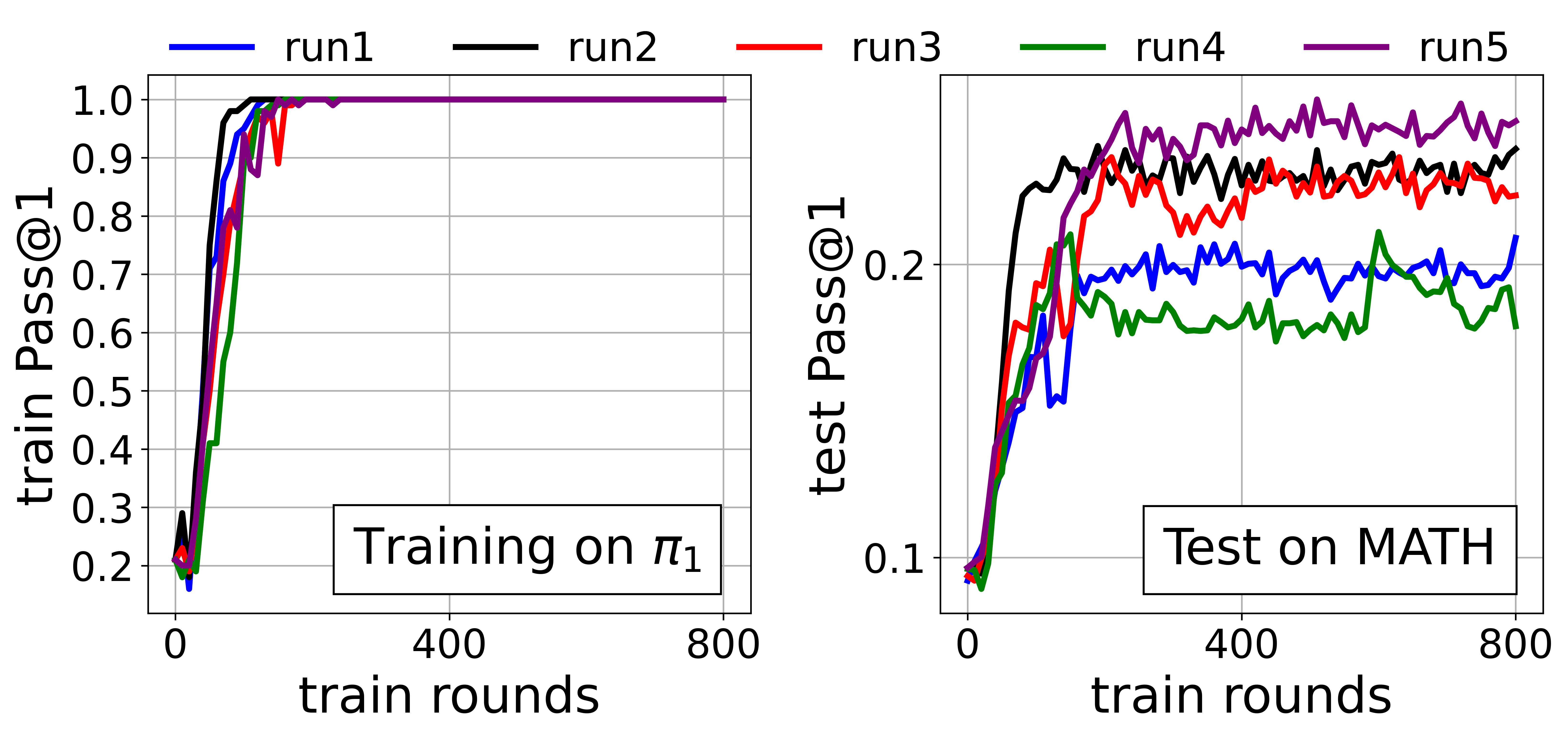}
  \label{only-reward-llama-pi1-Exp3}
}\\[-0.08in]
\subfigure[Qwen with 4 rollouts]{
  \includegraphics[width=0.95\linewidth]{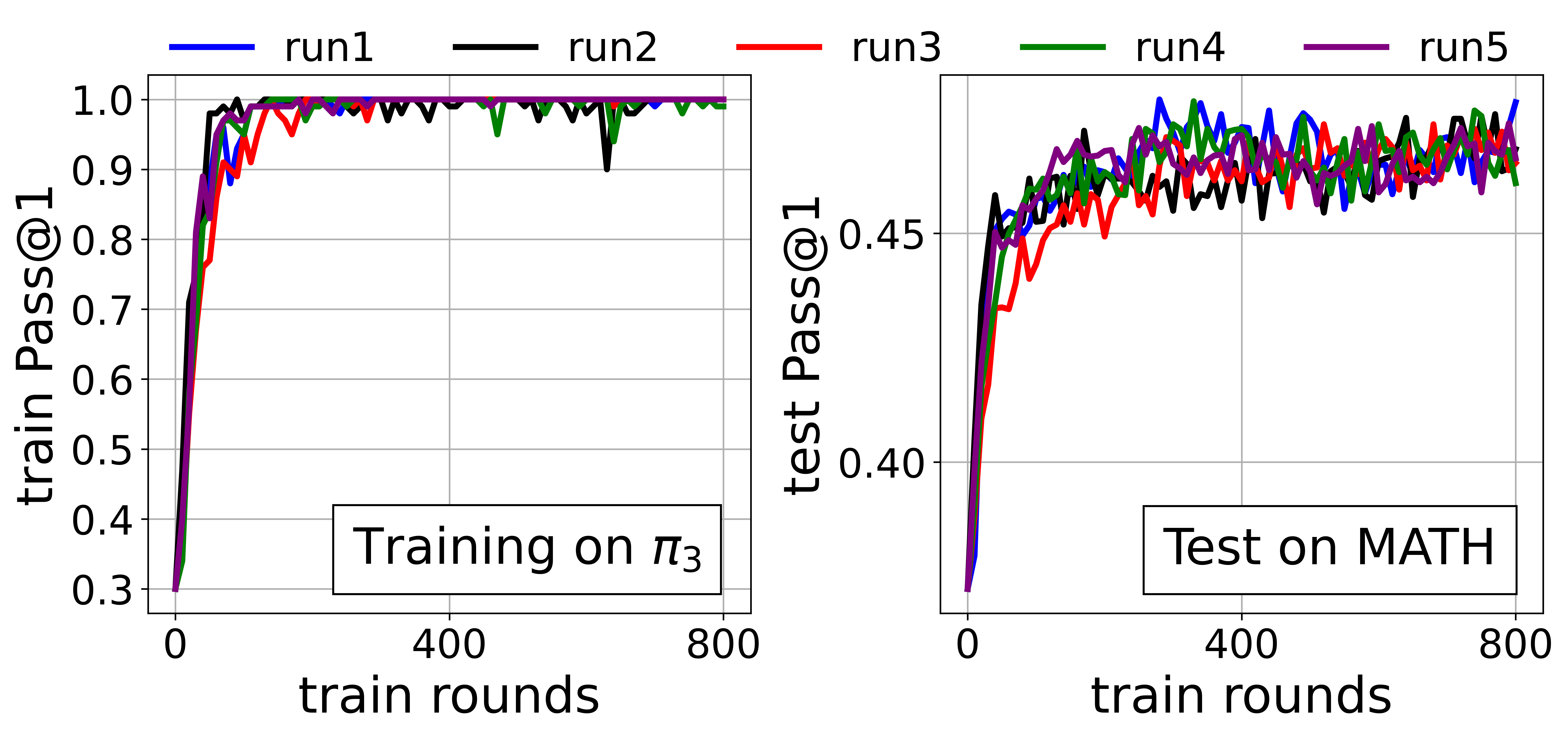}
  \label{only-reward-qwen15-pi3-Exp3}
}
\caption{Impact of number of rollouts (Math dataset).}
\label{fig:ImpactRollout-hard}
\end{minipage}

\vspace{-0.08in}
\end{figure}

\paragraph{Training dynamics.}  
The left columns of Figure \ref{fig:ImpactAdvantage-hard} show five training dynamics curves
of LLaMA and Qwen with the same training data in 
Figure \ref{fig:LLaMA-minimalist-hard} and \ref{fig:QWEN-minimalist-hard}.  
Comparing the left columns of Figure \ref{only-one-llama-pi1-Exp2} 
and Figure \ref{rollout-1-llama-hard-pi1-Exp1}, 
one can observe that with the GRPO advantage function, 
the variation in training dynamics curves of LLaMA becomes larger.  
Similar observations can be made on Qwen when comparing 
left columns of Figure \ref{only-one-qwen15-pi3-Exp2} and Figure \ref{rollout-1-qwen15-hard-pi3-Exp1}.  
This implies that the advantage function does not make the training more stable 
as claimed in previous works.  
One reason for this inconsistency is that  
in previous works the advantage function is usually coupled with multiple rollouts, 
which makes the performance gain contribution of advantage function less clear.  
This observation is not surprising, 
since the outcome reward is deterministic and it identifies the optimal action 
making the advantage function less critical in training.  

\paragraph{Generalizatoin dynamics.}  
The right columns of Figure \ref{fig:ImpactAdvantage-hard} show five generalization dynamics curves
of LLaMA and Qwen with the same training data in 
Figure \ref{fig:LLaMA-minimalist-hard} and \ref{fig:QWEN-minimalist-hard}.  
Comparing the right columns of Figure \ref{only-one-llama-pi1-Exp2}, Figure \ref{rollout-1-llama-hard-pi1-Exp1} and Figure \ref{only-one-qwen15-pi3-Exp2} and Figure \ref{rollout-1-qwen15-hard-pi3-Exp1}, 
one can observe that with GRPO advantage function, 
the generalization curves of both LLaMA and Qwen fluctuate more 
and the Pass@1 evaluated on the MATH test data 
varies slightly.
This implies that the advantage function does not make the 
generalization more stable as claimed in previous works.  
The underlying reason is similar to that explaining the training dynamics inconsistency.  

\paragraph{Key insights.}  
Compared with the minimalist configuration without advantage design, 
the GRPO advantage function makes both training and generalization less stable 
and the Pass@1 evaluated on the test data is not improved.  

\subsection{Exp3: Impact of Number of Rollouts}
To study the impact of number of rollouts,  
we only modify one component of the configuration in Section \ref{sec:Exp1}, 
i.e., increasing the number of rollouts from one to four, without incorporating any advantage design.  

\paragraph{Training dynamics.}  
The left columns of Figure \ref{only-reward-llama-pi1-Exp3} 
show five training dynamics curves of LLaMA with four rollouts in each training round.  
As the number of training rounds increases 
from one to around 100, the Pass@1 evaluated on the training data 
increases from around 0.2 to around 1.  
Recall from Figure \ref{rollout-1-llama-hard-pi1-Exp1} that with one rollout per round, 
it takes around 200 rounds to converge.  
This implies a significant reduction in the number of training rounds needed to converge.  
From a sample efficiency perspective, though using half of rounds to converge, each round costs four times of rollouts, 
implying a drop in sample efficiency.  
Comparing Figure \ref{only-reward-qwen15-pi3-Exp3} with Figure \ref{rollout-1-qwen15-hard-pi3-Exp1}, 
one can observe a smaller reduction ratio in training rounds to converge, 
implying a higher drop in sample efficiency caused by 
increasing the number of rollouts.  
These observations are aligned with multi-armed bandit learning that pulling multiple arms improves the learning speed in terms of rounds but leads to sample efficiency drop.   

\paragraph{Generalization dynamics.}  
The right columns of Figure \ref{only-reward-llama-pi1-Exp3} 
show five generalization dynamics curves of LLaMA with four rollouts in each training round.  
The curves in Figure \ref{only-reward-llama-pi1-Exp3}  have roughly 
the same variation range as Figure \ref{rollout-1-llama-hard-pi1-Exp1} 
and the Pass@1 evaluated on the MATH testing dataset is roughly the same. Comparing the right columns of 
Figure \ref{only-reward-qwen15-pi3-Exp3} with Figure \ref{rollout-1-qwen15-hard-pi3-Exp1}, 
a similar pattern is observed on Qwen.  
It shows that though generating more rollouts during training 
improves learning speed, 
it does not improve 
the test Pass@1.  
This observation is quite different from 
previous works that suggest scaling the 
rollouts improves model performance.  
At first glance, one may find that another inconsistency arises.  
But after further thoughts, it implies that rollout scaling needs 
deeper understanding, and the current understanding is flawed.  

\paragraph{Key insights. } 
Increasing the number of rollouts improves learning speed but leads to a drop in sample efficiency.  
Furthermore, it does not improve generalization capability.    

\subsection{Exp4: Impact of Extremely Difficult Data}

\begin{figure}[!htb]
\centering
\vspace{-0.08in}

% -------- Figure A: LLaMA --------
\begin{minipage}[t]{0.49\linewidth}
\centering
\subfigure[LLaMA with Pass@1 = 0.03]{
  \includegraphics[width=0.95\linewidth]{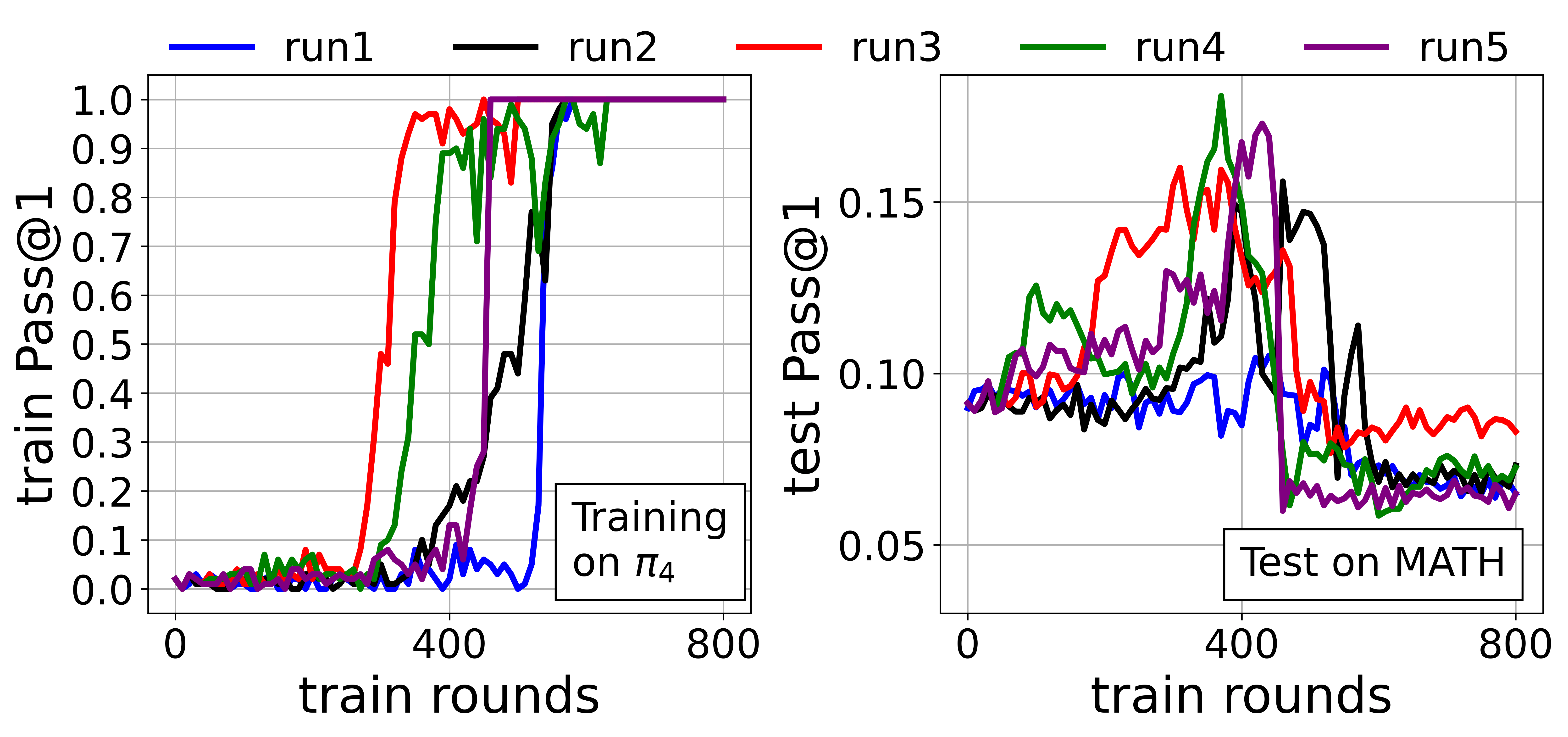}
  \label{rollout-1-llama-pi4-Exp4}
}\\[-0.08in]
\subfigure[Qwen with Pass@1 = 0.02]{
  \includegraphics[width=0.95\linewidth]{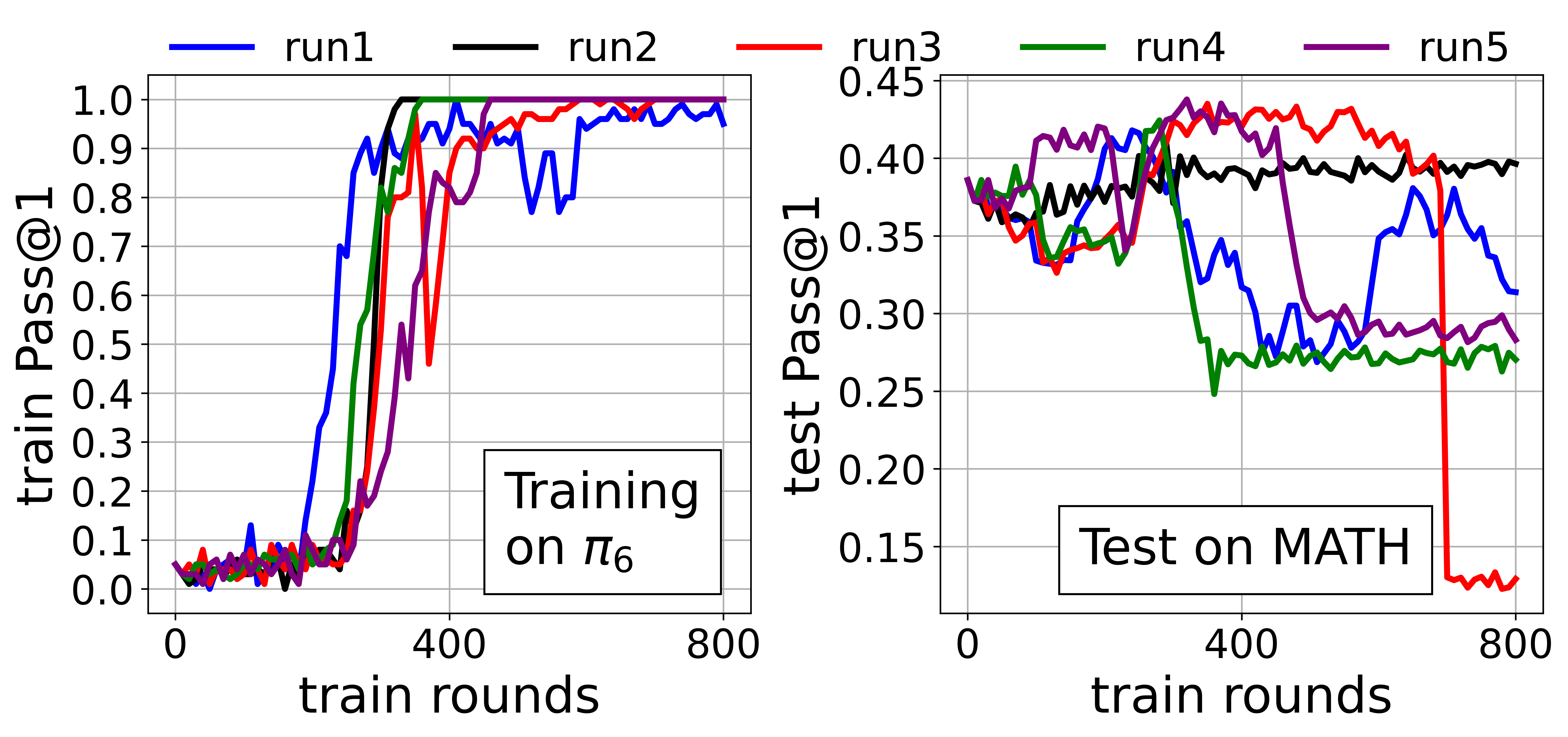}
  \label{rollout-1-qwen15-pi6-Exp4}
}
\caption{Impact of extremely difficult data (without advantage).}
\label{fig:ImpactLongTail-hard-without}
\end{minipage}
\hfill
% -------- Figure B: Qwen --------
\begin{minipage}[t]{0.49\linewidth}
\centering
\subfigure[LLaMA with Pass@1 = 0.03]{
  \includegraphics[width=0.95\linewidth]{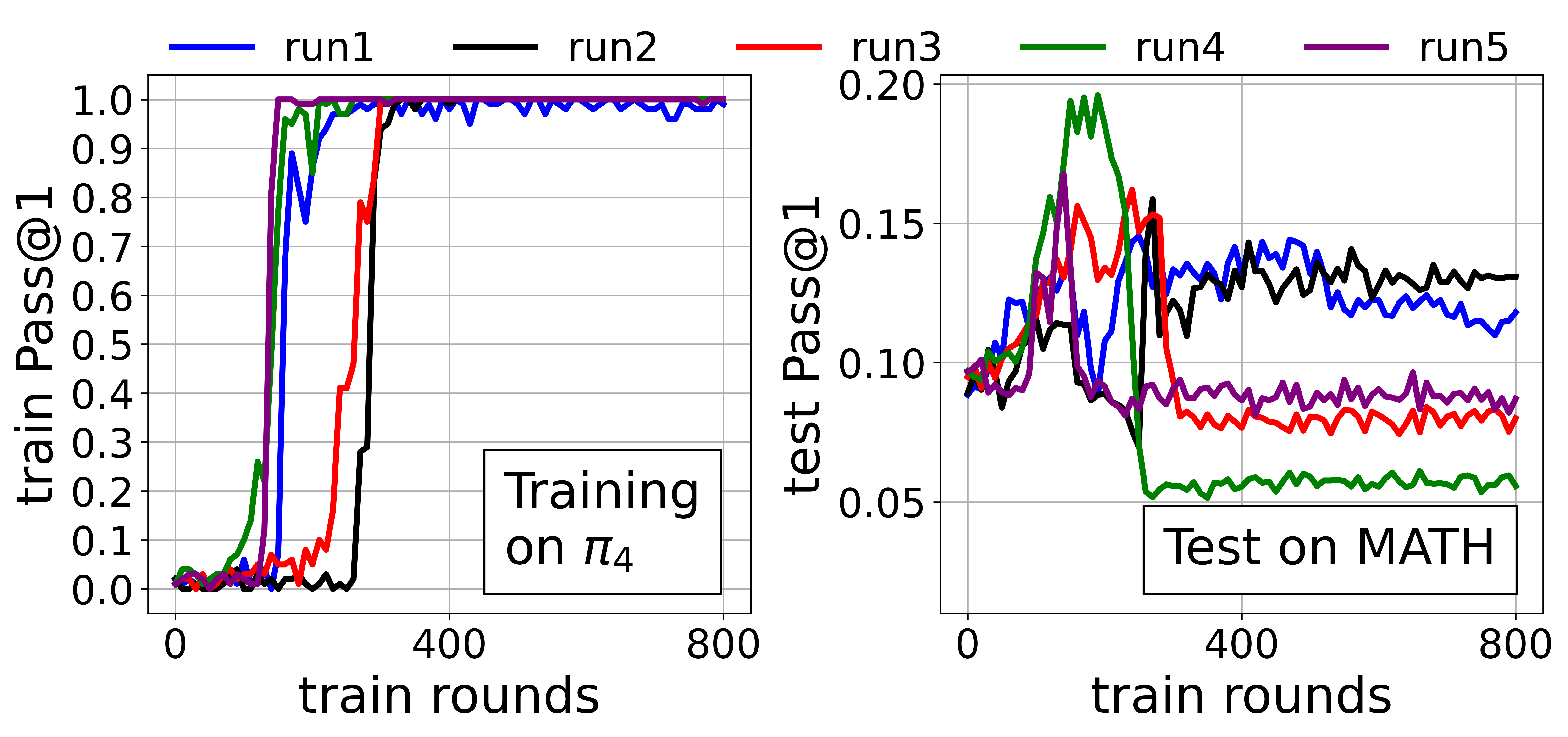}
  \label{only-one-llama-pi4-Exp4}
}\\[-0.08in]
\subfigure[Qwen with Pass@1 = 0.02]{
  \includegraphics[width=0.95\linewidth]{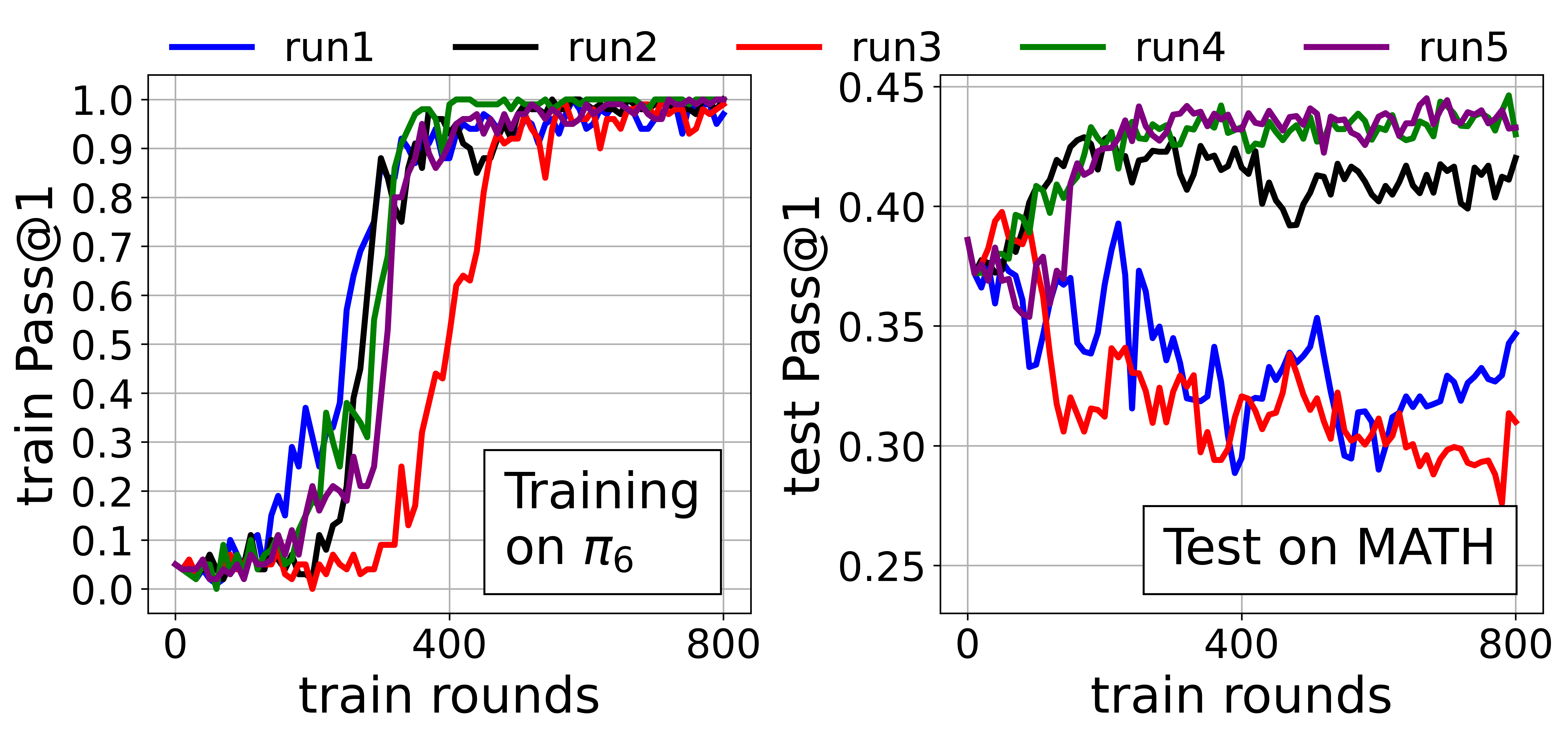}
  \label{only-one-qwen15-pi6-Exp4}
}
\caption{Impact of extremely difficult data (with advantage).}
\label{fig:ImpactLongTail-hard-with}
\end{minipage}

\vspace{-0.08in}
\end{figure}

Our results thus far studied training data with moderate difficulty, 
i.e., the base model has Pass@1 ranging from 0.1 to 0.8. 
We investigate the extreme case of 
training data with very low Pass@1.  
To achieve this, we only modify one component of the configuration in Section \ref{sec:Exp1}, 
i.e., replacing the training data with one that the base model has  
Pass@1 below 0.05.  
Eventually, we selected different training data for LLaMA and Qwen, 
since they have different capabilities and the extremely difficult data instances for 
them are different.   

\paragraph{Training dynamics.} 
The left columns of 
Figure \ref{fig:ImpactLongTail-hard-without} show the training dynamics of LLaMA 
and Qwen with extremely difficult training data.  
The curves follow a similar pattern 
to that in Figure \ref{rollout-1-llama-hard-pi1-Exp1}, but it takes around 400 rounds to converge, compared to 200 rounds in Figure \ref{rollout-1-llama-hard-pi1-Exp1}.  
This implies a drastic drop in learning speed.  
Similar observations hold for Qwen as indicated by 
the left columns of Figure \ref{rollout-1-qwen15-pi6-Exp4} and 
Figure \ref{rollout-1-qwen15-hard-pi3-Exp1}.  
This implies that under the minimalist configuration, 
the optimal policy for the extremely difficult data can still be learned.  
The left columns of Figure \ref{fig:ImpactLongTail-hard-with} incorporate 
the advantage function following the same setting as Section \ref{sec:advF}.  
One can observe that for LLaMA, the number of rounds to converge is shortened from 
around 400 to around 250, a significant reduction.  
However, for Qwen the number of rounds to converge roughly remains the same.    

\paragraph{Generalization dynamics.}  
The right columns of 
Figure \ref{fig:ImpactLongTail-hard-without} show the training dynamics of LLaMA 
and Qwen with extremely difficult training data.  
It shows that when the training rounds increase, 
the Pass@1 evaluated on the test data first increases, 
and then suffers a drastic drop.  
After the drop, the Pass@1 is even smaller 
than that of the base model, implying 
that the model is experiencing catastrophic forgetting.  
It is interesting to observe that in most cases, 
the drop 
happens when the learning hits Pass@1 with 1. 
This also implies that the extremely difficult data may hurt the generalization.  
The right columns of Figure \ref{fig:ImpactLongTail-hard-with} incorporate 
the advantage function following the same setting as Section \ref{sec:advF}.  
One can observe that by incorporating advantage function, 
two generalization dynamics curves for LLaMA and three for Qwen 
do not suffer from a drastic drop anymore, 
though the remaining curves still suffer.  
This implies that the advantage function reduces the risk of catastrophic forgetting 
to a certain degree when the model is fine-tuned with extremely difficult data. 

\paragraph{Key insights.} 
When the training data is extremely difficult (i.e., Pass@1 $<$ 0.05), 
under the minimalist configuration, the model can still learn the optimal policy, 
but with more learning rounds.  
But the Pass@1 over the testing data suffers drastic drop (even below base model), 
incorporating advantage function can reduce this risk to a certain degree.  

\subsection{Exp5: Impact of Reward Design}
We consider the same setting as Section \ref{sec:Exp1}, except that we 
replace the reward metric $\{0,1\}$ with $\{-1,0\}$.  

\begin{figure}[!htb]
\centering
\vspace{-0.08in}

% -------- Figure A: LLaMA --------
\begin{minipage}[t]{0.49\linewidth}
\centering
\subfigure[LLaMA with reward $\{-1,0\}$]{
  \includegraphics[width=0.95\linewidth]{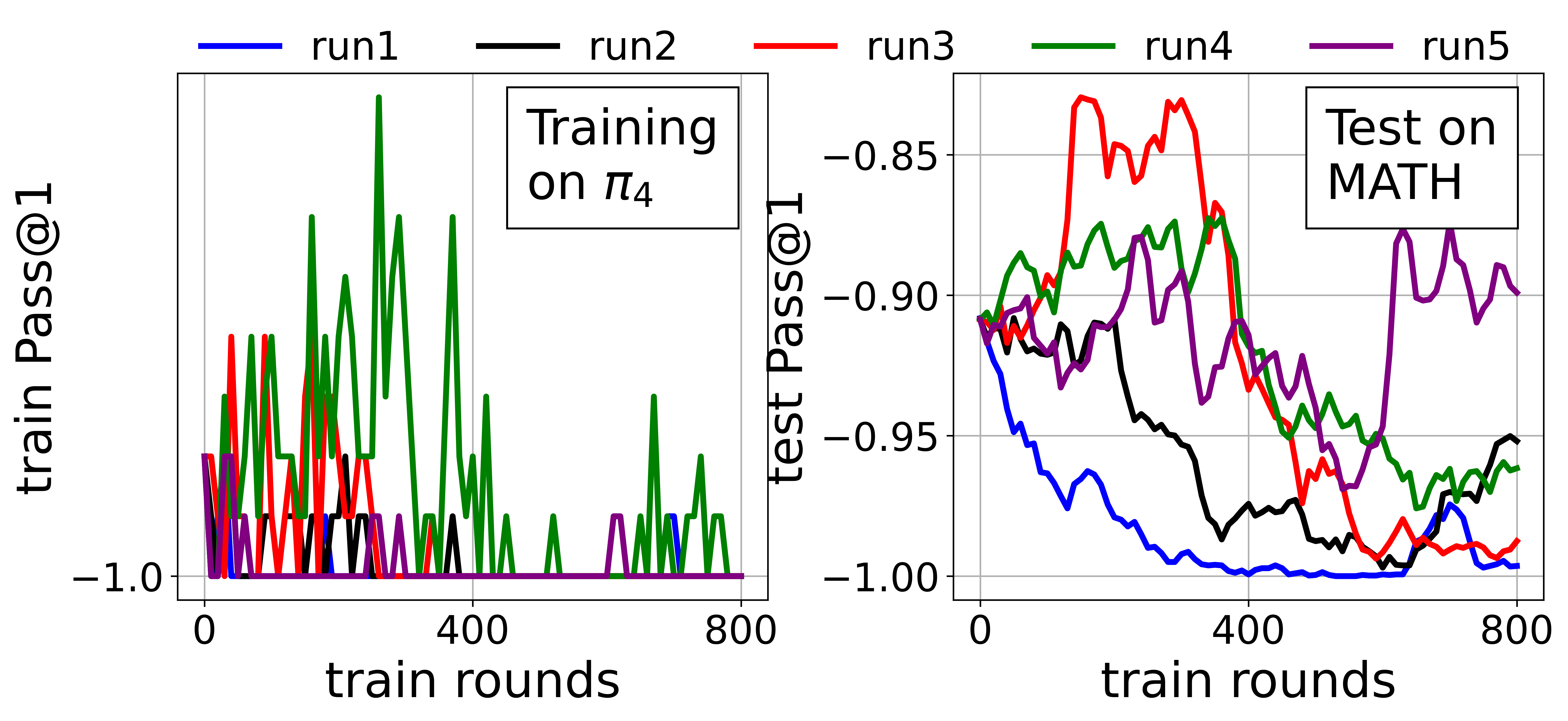}
  \label{neg-reward-llama-pi4-Exp5}
}\\[-0.08in]
\subfigure[Qwen with reward $\{-1,0\}$]{
  \includegraphics[width=0.95\linewidth]{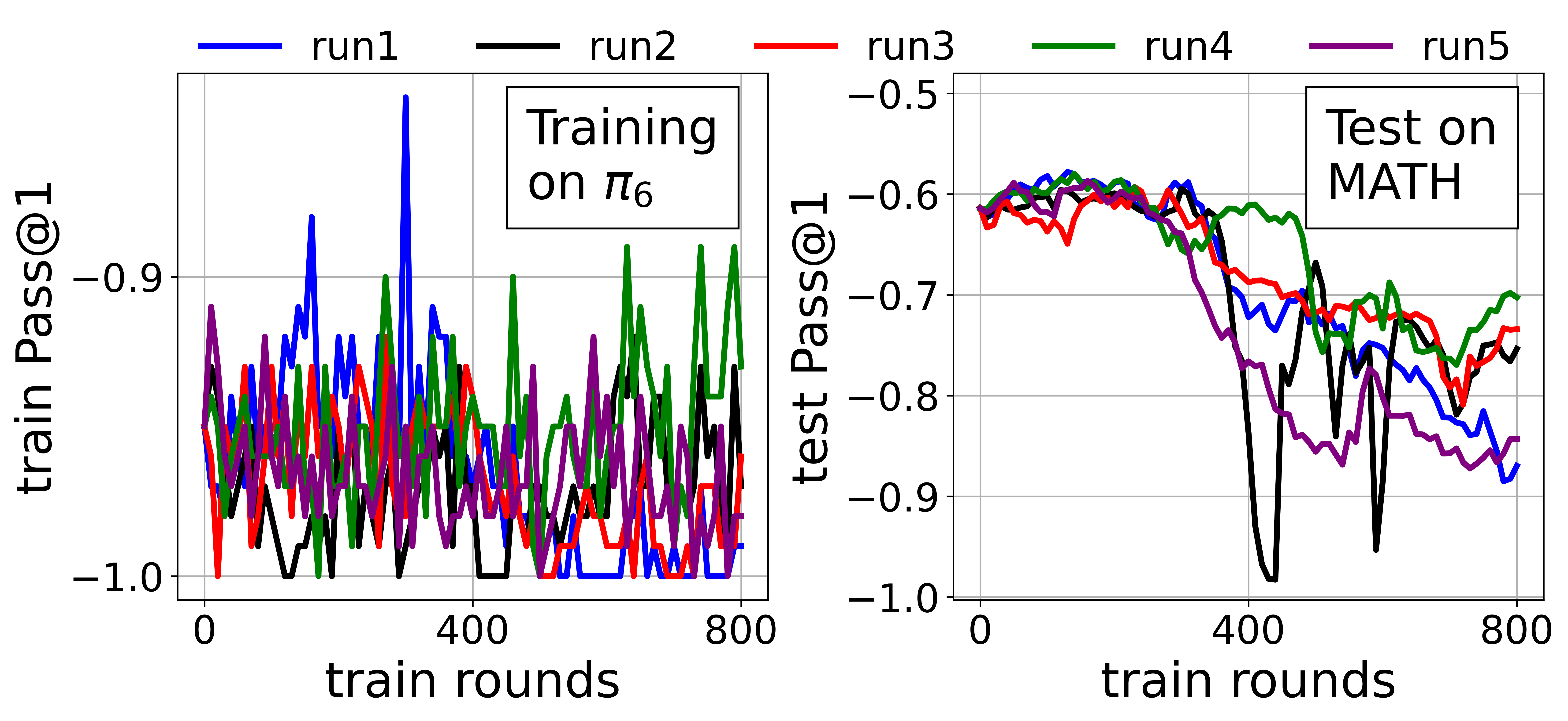}
  \label{neg-reward-qwen15-pi6-Exp5}
}
\caption{Impact of reward design.}
\label{fig:ImpactRewardDesign-hard}
\end{minipage}
\hfill
% -------- Figure B: Qwen --------
\begin{minipage}[t]{0.49\linewidth}
\centering
\subfigure[OLMo with difficult training data]{
  \includegraphics[width=0.95\linewidth]{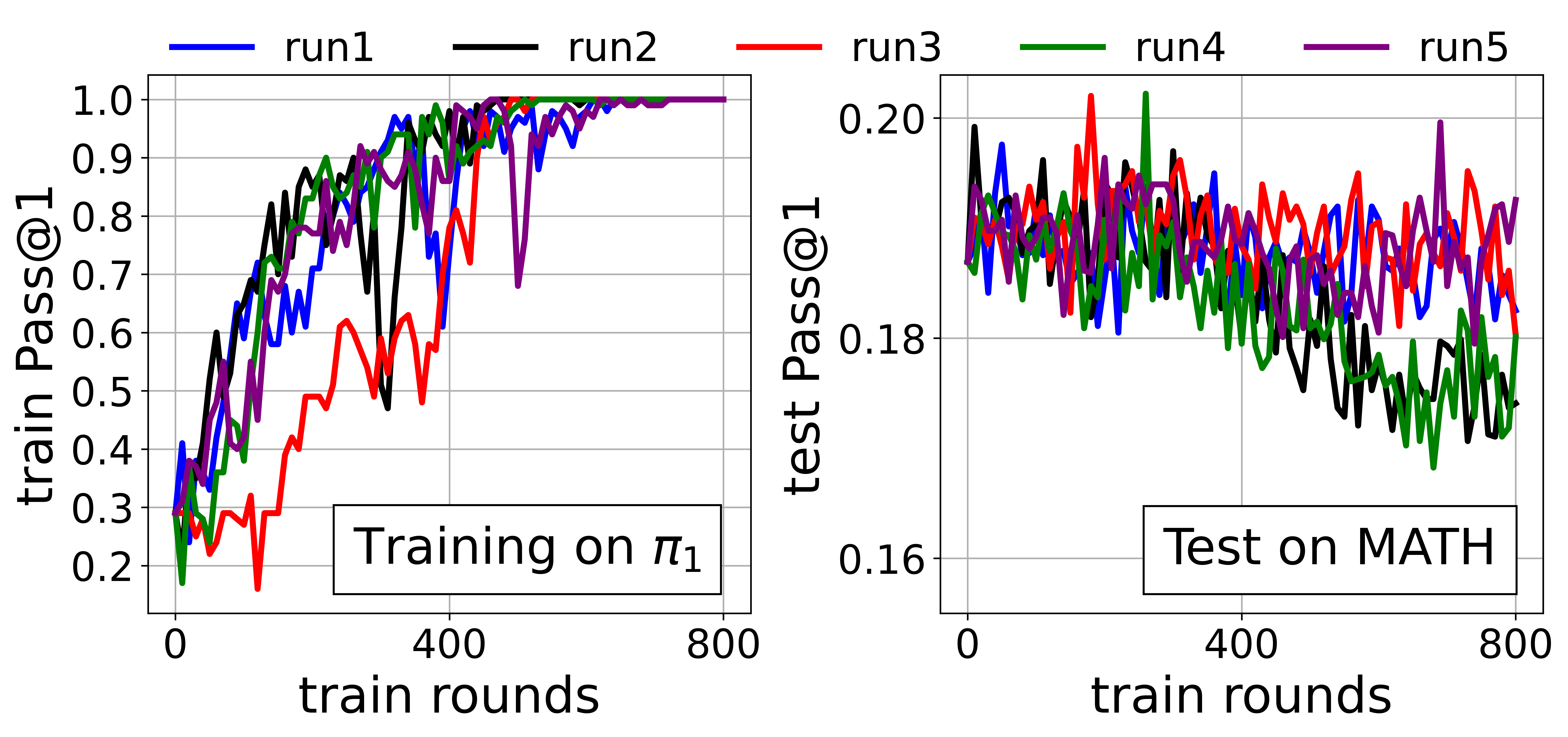}
  \label{rollout-1-olmo-hard-pi1-Exp6}
}\\[-0.08in]
\subfigure[OLMo with simple training data]{
  \includegraphics[width=0.95\linewidth]{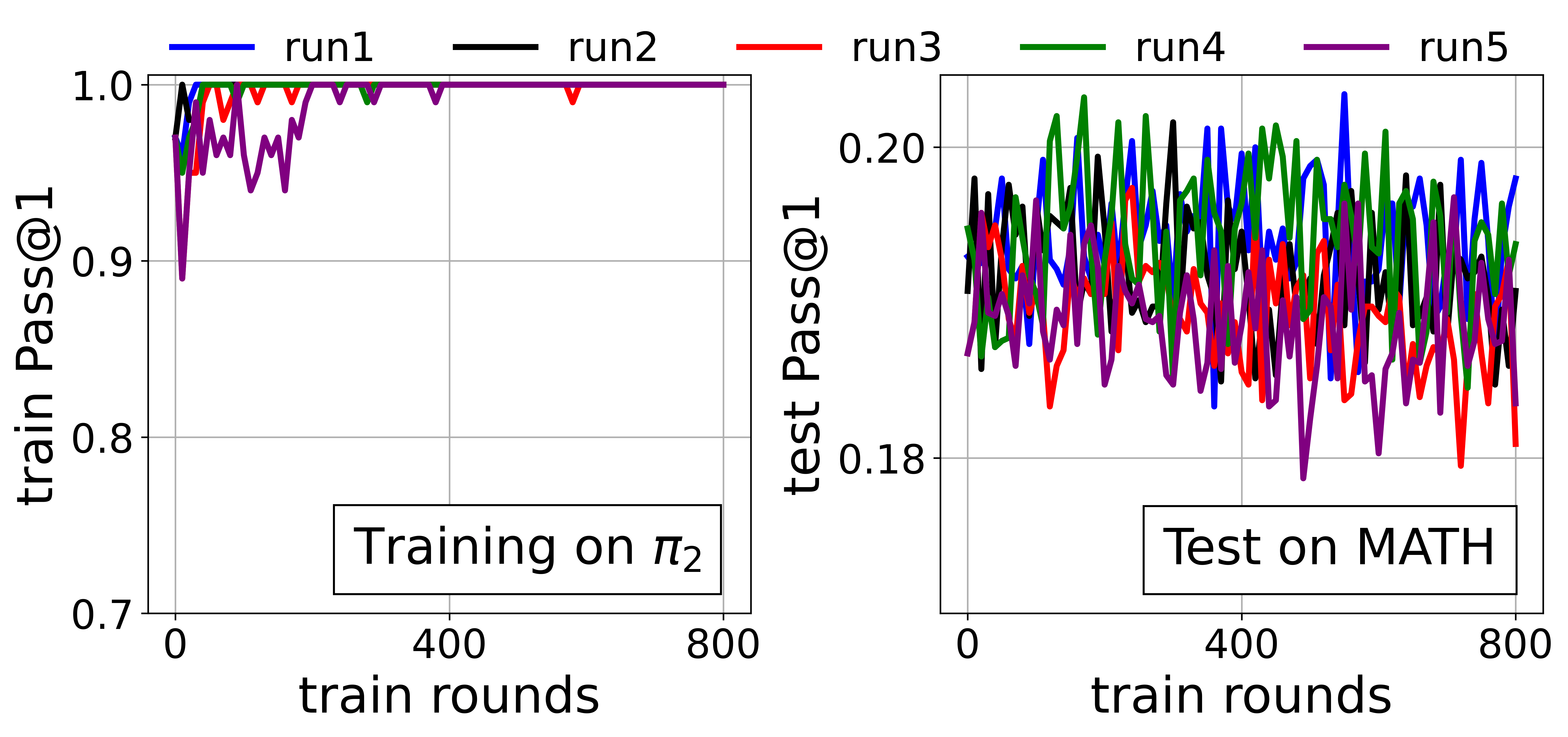}
  \label{rollout-1-olmo-easy-pi2-Exp6}
}
\caption{Impact of base model.}
\label{fig:ImpactBaseModel-hard}
\end{minipage}

\vspace{-0.08in}
\end{figure}

\paragraph{Learning dynamics.} 
The left columns of Figure \ref{fig:ImpactRewardDesign-hard} show the training 
dynamics curves of LLaMA and Qwen.  
One can observe that these curves are highly fluctuating 
and they concentrate around -1, far from zero.  
This implies that under the negative reward function, 
the model fails to learn the optimal policy.  
The traditional multi-armed bandit learning literature is almost 
invariant to reward metric shift. 

\paragraph{Generalization dynamics.}  
The right columns of Figure \ref{fig:ImpactRewardDesign-hard} show the generalization 
dynamics curves of LLaMA and Qwen.  
Almost all curves have a trend of continuously decreasing.  
In other words, the model does not generalize. 
This implies that fine-tuning with negative rewards may hurt the model's capability.  

\paragraph{Key insights. }
The negative reward metric leads the model to fail to learn the optimal policy, 
and it fails to generalize.  

\subsection{Exp6: Impact of Base Model}
Lastly, we report experiment results on OLMo, 
which reveal deeper insights into the impact of base model.  
We still consider the minimalist configuration in Section \ref{sec:Exp1}.  
 
\paragraph{Training dynamics.}  
The left columns of Figure \ref{fig:ImpactBaseModel-hard} show 
five training dynamics curves for OLMo.  
One can observe that no matter whetherthe training data is difficult or easy, 
the learning curves have a similar pattern to LLaMA and Qwen.  
This means that the OLMo still can learn the optimal policy.  
 
\paragraph{Generalization dynamics.}  
The right columns of Figure \ref{fig:ImpactBaseModel-hard} show 
five generalization dynamics curves for OLMo.  
One can observe that all curves are highly fluctuating like random noise.  
Furthermore, each curve does not exhibit an increasing trend, 
and some even show a decreasing trend.  
This shows that the generalization capability is not improved at all.  
This shows that OLMo is quite different from LLaMA and Qwen.  
 
\paragraph{Key insights.}  
The OLMo has similar training dynamics to LLaMA and Qwen. 
Though it learns the optimal policy for data with different 
difficulty levels, it does not generalize.

\section{Conclusion}  
We designed an extremely simple comparison baseline: 
one training data, one rollout per round and 
the reward serving as the advantage.  
Drawing critical connections  to 
multi-armed bandit learning 
with extremely large discrete action space, 
we designed a generic experimental pipeline 
to carefully examine the role of each design choice. 
Experimental results on three LLMs 
and two datasets not only reveal new understanding 
of the design choice but also identify important bottlenecks 
that the reinforcement fine-tuning community should pay more attention to. 

\appendix

\bibliographystyle{named}
\bibliography{references}

\section*{Supplementary Overview}
This supplementary material provides additional experimental details and results that support the main findings of the paper. Specifically,

(i) Section A contains further details of our experimental setup.

(ii) Section B lists the exact training and evaluation problems used in our experiments

\section{Experiment Setting}

\paragraph{Base Models.}
We consider three instruction-tuned large language models: LLaMA-3.2-1B-Instruct, OLMo-2-0425-1B-Instruct, and Qwen2.5-1.5B-Instruct. These models represent diverse pretraining and alignment pipelines and are commonly used as lightweight backbones for reinforcement fine-tuning.

\paragraph{Datasets.}
We adopt two open-source mathematical reasoning benchmarks, i.e., 
MATH and GSM8K, for training and evaluation. Following a single-example training protocol, each training dataset consists of a single problem instance sampled from either MATH or GSM8K. To study the effect of problem difficulty, we construct multiple such single-example training sets by selecting problems with varying difficulty levels. 
Specifically, the selected MATH problems are denoted as $\pi_1, \pi_2, \ldots, \pi_7$, and the selected GSM8K problems are denoted as $\phi_1$. The exact training problems and their ground-truth answers are provided in appendix.  

\paragraph{Evaluation.}
We evaluate model performance using the Pass@1 metric, 
which measures the fraction of test problems for which the model produces a correct solution
in its first generated response. Evaluation is performed on the official test split of the corresponding benchmark (MATH test or GSM8K test). During evaluation, responses are generated using the same sampling parameters as those used during training, with a temperature of 1.0 and top-$p$ of 1.0, corresponding to standard softmax sampling. 
For evaluation on the training set, results are computed by averaging over 100 independently generated responses for training problem, which serves to monitor training dynamics during optimization.  

\paragraph{Base Parameter Settings.}
We employ the VeRL framework to perform GRPO training. Unless otherwise specified, we use the following hyperparameters for all experiments: the KL-divergence coefficient is set to 0.001, the PPO clipping ratio is 0.2, and the learning rate is $1\times10^{-6}$. The maximum prompt length and response length are set to 512 and 1024 tokens, respectively. To match the single-task-new training setting, we set both the batch size and mini-batch size to 1.

\paragraph{Hardware.}
All experiments are conducted using a single GPU. Specifically, LLaMA-3.2-1B-Instruct and OLMo-2-0425-1B-Instruct are trained on a single A100 GPU(40GB), while Qwen2.5-1.5B-Instruct is trained on a single H800 GPU(80GB).

\section{Detailed Problem List for Reproducibility}
To facilitate reproducibility and transparent analysis, we explicitly list the problems used in our experiments. The problems are selected based on their difficulty, which is quantified by the accuracy of a fixed base model
prior to any reinforcement fine-tuning.

Concretely, for each problem we generate 100 single responses from the base model and compute the empirical accuracy as the fraction of correct responses. We then select problems according to predefined difficulty thresholds (e.g., accuracy $<0.05$ for extremely difficult cases) prior to any reinforcement fine-tuning.

$\pi_1$
\begin{tcolorbox}[
    colback=white,
    colframe=black,
    boxrule=0.5pt,
    left=6pt,
    right=6pt,
    top=4pt,
    bottom=4pt
]
\textbf{Problem:} When the expression $(2x^4+3x^3+x-14)(3x^{10}-9x^7+9x^4+30)-(x^2+5)^7$ is expanded, what is the degree of the resulting polynomial?

\textbf{Ground truth:} 14

\textbf{Difficulty:} Level 4
\end{tcolorbox}

$\pi_2$

\begin{tcolorbox}[
    colback=white,
    colframe=black,
    boxrule=0.5pt,
    left=6pt,
    right=6pt,
    top=4pt,
    bottom=4pt
]
\textbf{Problem:} Evaluate: $64^2-36^2$ 

\textbf{Ground truth:} 2800

\textbf{Difficulty:} Level 1
\end{tcolorbox}

$\pi_3$
\begin{tcolorbox}[
    colback=white,
    colframe=black,
    boxrule=0.5pt,
    left=6pt,
    right=6pt,
    top=4pt,
    bottom=4pt
]
\textbf{Problem:} Convert the point $(0, -3 \sqrt{3}, 3)$ in rectangular coordinates to spherical coordinates.  Enter your answer in the form $(\rho,\theta,\phi),$ where $\rho > 0,$ $0 \le \theta < 2 \pi,$ and $0 \le \phi \le \pi.$ 

\textbf{Ground truth:} $\left( 6, \frac{3 \pi}{2}, \frac{\pi}{3} \right)$

\textbf{Difficulty:} Level 4
\end{tcolorbox}

$\pi_4$
\begin{tcolorbox}[
    colback=white,
    colframe=black,
    boxrule=0.5pt,
    left=6pt,
    right=6pt,
    top=4pt,
    bottom=4pt
]
\textbf{Problem:} For how many values of $a$ is it true that:

(1) $a$ is a positive integer such that $a \le 50$.

(2) the quadratic equation $x^2 + (2a+1)x + a^2 = 0$ has two integer solutions? 

\textbf{Ground truth:} 6

\textbf{Difficulty:} Level 5
\end{tcolorbox}

$\pi_5$
\begin{tcolorbox}[
    colback=white,
    colframe=black,
    boxrule=0.5pt,
    left=6pt,
    right=6pt,
    top=4pt,
    bottom=4pt
]
\textbf{Problem:} Let $f(x) = 2x - 3$ and $g(x) = x + 1$. What is the value of $f(1 + g(2))$? 

\textbf{Ground truth:} 5

\textbf{Difficulty:} Level 3
\end{tcolorbox}

$\pi_6$
\begin{tcolorbox}[
    colback=white,
    colframe=black,
    boxrule=0.5pt,
    left=6pt,
    right=6pt,
    top=4pt,
    bottom=4pt
]
\textbf{Problem:} Let $\omega$ be a nonreal root of $z^3 = 1.$  Let $a_1,$ $a_2,$ $\dots,$ $a_n$ be real numbers such that
\[\frac{1}{a_1 + \omega} + \frac{1}{a_2 + \omega} + \dots + \frac{1}{a_n + \omega} = 2 + 5i.\]Compute
\[\frac{2a_1 - 1}{a_1^2 - a_1 + 1} + \frac{2a_2 - 1}{a_2^2 - a_2 + 1} + \dots + \frac{2a_n - 1}{a_n^2 - a_n + 1}.\] 

\textbf{Ground truth:} 4

\textbf{Difficulty:} Level 5
\end{tcolorbox}

$\pi_7$
\begin{tcolorbox}[
    colback=white,
    colframe=black,
    boxrule=0.5pt,
    left=6pt,
    right=6pt,
    top=4pt,
    bottom=4pt
]
\textbf{Problem:} Given that $n$ is an integer and $0 < 4n <30$, what is the sum of all possible integer values of $n$?

\textbf{Ground truth:} 28

\textbf{Difficulty:} Level 2
\end{tcolorbox}

$\phi_1$
\begin{tcolorbox}[
    colback=white,
    colframe=black,
    boxrule=0.5pt,
    left=6pt,
    right=6pt,
    top=4pt,
    bottom=4pt
]
\textbf{Problem:} The school nurse must conduct lice checks at the elementary school. She must check 26 Kindergarteners, 19 first graders, 20 second graders, and 25 third graders. If each check takes 2 minutes, how many hours will it take the nurse to complete all the checks? 

\textbf{Ground truth:} 3
\end{tcolorbox}

% \paragraph{$\phi_2$}
% Grayson drives a motorboat for 1 hour at 25 mph and then 0.5 hours for 20 mph. Rudy rows in his rowboat for 3 hours at 10 mph. How much farther, in miles, does Grayson go in his motorboat compared to Rudy? 

% ground truth:
% 5

\end{document}